\documentclass[letterpaper, 10 pt, conference]{ieeeconf}  

\IEEEoverridecommandlockouts                              

\usepackage{graphicx}
\usepackage{booktabs}
\usepackage{threeparttable}
\usepackage{multirow}
\usepackage{array}
\usepackage{makecell}
\usepackage{svg}
\usepackage{amsfonts}
\usepackage{amsmath}
\usepackage{bbm}
\usepackage{algorithm}
\usepackage{algorithmic}
\usepackage{subcaption}
\usepackage{cite}
\usepackage{dsfont}

\title{\LARGE \bf
Learning Terrain-Specialized Policies for Adaptive Locomotion in Challenging Environments
}

\author{Matheus P. Angarola$^{*}$, Francisco Affonso$^{*}$ and Marcelo Becker
\thanks{$*$These authors contributed equally.}
\thanks{The publication was written prior to Francisco Affonso joining University of Illinois Urbana Champaign (UIUC).}
\thanks{All authors are with the Department of Mechanical Engineering, University of São Paulo (USP), BR {\tt\small \{matheuzu11, faffonso\}@usp.br, becker@sc.usp.br}}%
}

\definecolor{pastelblue}{RGB}{30,80,160}

\begin{document}

\maketitle
\thispagestyle{empty}
\pagestyle{empty}

\begin{abstract}

Legged robots must exhibit robust and agile locomotion across diverse, unstructured terrains, a challenge exacerbated under blind locomotion settings where terrain information is unavailable. This work introduces a hierarchical reinforcement learning framework that leverages terrain-specialized policies and curriculum learning to enhance agility and tracking performance in complex environments. We validated our method on simulation, where our approach outperforms a generalist policy by up to 16\% in success rate and achieves lower tracking errors as the velocity target increases, particularly on low-friction and discontinuous terrains, demonstrating superior adaptability and robustness across mixed-terrain scenarios.

\end{abstract}


\section{Introduction}

Legged robots operating in real-world environments must function reliably across complex and unstructured terrains to achieve task-oriented goals, requiring locomotion controllers that are both agile and adaptable. Traditional approaches based on model predictive control (MPC) and trajectory optimization (TO) have demonstrated effectiveness under nominal conditions but often fail in edge cases, where simplified dynamics and constraint models are insufficient to capture the complexities of robot–environment interactions~\cite{lu2023whole, grandia2023perceptive}.

To address these limitations, learning-based methods have seen growing adoption, leveraging the capacity of neural networks to model high-dimensional, nonlinear control problems from data~\cite{choi2023learning}. In particular, deep reinforcement learning (RL) has emerged as a promising paradigm, enabling agents to learn control policies through direct interaction with the environment, without requiring curated datasets---a fundamental bottleneck in supervised learning~\cite{affonso2025learningwalklessdynastyle, yang2020data}.

In RL-based locomotion, control is formulated as a sequential decision-making problem, where policies are optimized through trial-and-error to maximize expected cumulative rewards, typically using simulation environments~\cite{rudin2022learning}. To represent the diversity of real-world conditions, domain randomization is applied to factors such as terrain profiles, external perturbations, and sensor noise. Legged robots can learn robust mappings from sensory observations to control actions, enabling them to track desired velocity commands while maintaining stability across diverse terrains~\cite{Katz23, haarnoja2019learning, song2025learning}.

\begin{figure}[h]
    \centering
    \includegraphics[width=1\linewidth]{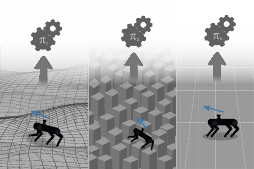}
    \caption{Hierarchical locomotion control architecture based on terrain-specialized policies. The system selects an appropriate expert policy based on the perceived terrain to execute the desired locomotion behavior. The blue arrow \textcolor{pastelblue}{$(\rightarrow)$} represents the robot’s velocity direction.}
    \label{fig:initial_image}
\end{figure}

Furthermore, a particular class of locomotion frameworks, known as blind policies, relies solely on proprioceptive information, without access to exteroceptive sensors such as cameras or LiDAR. These policies are typically employed in systems with limited sensing capabilities or constrained computational resources. However, the lack of terrain awareness encourages the controller to operate under worst-case assumptions, as the robot can only sense the terrain after physical contact rather than perceive it in advance. This reduces agility and limits grades overall locomotion performance, especially when tracking velocity commands in challenging environments~\cite{luo2017robust, jin2022high}.

In parallel, prior work has demonstrated the effectiveness of hierarchical learning in solving complex locomotion and navigation tasks, where a high-level policy selects among low-level behaviors, each specialized for a specific context~\cite{li2020learning, hoeller2024anymal}. Nevertheless, this framework remains underexplored in the context of adaptive locomotion with terrain-specialized policies, particularly in blind locomotion settings.

In this paper, we present a hierarchical RL framework for blind legged locomotion that leverages terrain-specialized control policies (Fig.~\ref{fig:initial_image}). Our method decomposes the locomotion task into terrain-specific subtasks, each addressed by a dedicated low-level policy trained with proprioceptive inputs and tailored reward shaping. A high-level policy selector uses privileged observations during deployment to identify the terrain type and activate the corresponding expert policy. To enable agile behavior, each policy is trained under a progressive curriculum that gradually expands the range of velocity commands based on performance.

Through systematic experimentation with terrain-specialized policies, we analyze their performance at high speeds, achieved by curriculum learning strategy, in complex environments. Furthermore, we compare the effectiveness of specialized versus generalized policies under equivalent training conditions, highlighting trade-offs in robustness and adaptability.

The key contributions of this work are:
\begin{itemize}
    \item Hierarchical control with terrain-specialized policies for blind locomotion
    \item Curriculum learning tailored to terrain-specific agility
\end{itemize}
\section{Related Work}

\textit{RL-Based Locomotion} can generate control actions without relying on explicit kinodynamic models or contact constraint formulations. Instead, the policy learns a direct mapping from observations to actions, guided by reward functions that not only encourage command tracking but also penalize undesirable behaviors. This enables flexible policy without imposing restrictive motion patterns~\cite{lee2020learning, Hwangbo19}. 

Miki et al.~\cite{miki2022learning} and Acero et al.~\cite{acero2022learning} demonstrated that, when the robot has access to perceptive information the policy can generalize across a wide range of environments and conditions, provided it is exposed to sufficient variability during training (e.g., different terrains, sensor noise).

Nevertheless, in the context of blind locomotion, the inability to perceive the current or upcoming terrain necessitates alternative strategies to enrich the policy's input. Prior works have explored the use of observation histories to implicitly encode information not directly available to the robot, thereby providing additional context for action selection. For instance, Kumar et al.~\cite{kumar2021rma} applied this idea to infer latent variables such as terrain friction and robot mass. Additionally, Margolis et al.~\cite{margolis2023walk} incorporated memory-based network architectures to estimate unobserved proprioceptive signals (e.g., body velocity), improving control performance under sensing constraints.

While these approaches offer valuable improvements to blind locomotion, they do not explicitly explore decomposing the task into terrain-specialized policies. Such a strategy could simplify the overall problem by partitioning it into more tractable subtasks and potentially improve performance in edge cases, particularly those requiring greater agility.

\textit{Hierarchical learning} is a framework that enables the training of specialized policies to solve complex tasks by delegating decision-making to a high-level policy that selects among these options. Hoeller et al.~\cite{hoeller2024anymal} applied this concept to achieve agile navigation for quadrupedal robots by decomposing the locomotion task into distinct modules for complex parkour behaviors such as walking, climbing, and jumping. This modular structure allowed the robot to traverse challenging terrain more effectively. 

Building on this idea, our work aims to advance the limits of locomotion agility across diverse terrains by integrating terrain-specific specialization within a hierarchical learning framework. To support this, we revisit curriculum learning strategies that have previously demonstrated effectiveness and adapt them to expose the robot to progressively more challenging conditions during training~\cite{li2024learning}. For instance, Margolis et al.~\cite{margolis2024rapid} introduced an adaptive curriculum in which the difficulty of command tracking was gradually increased, leading to enhanced agility and higher speeds.
\begin{figure*}
    \centering
    \includegraphics*[width=1.0\linewidth]{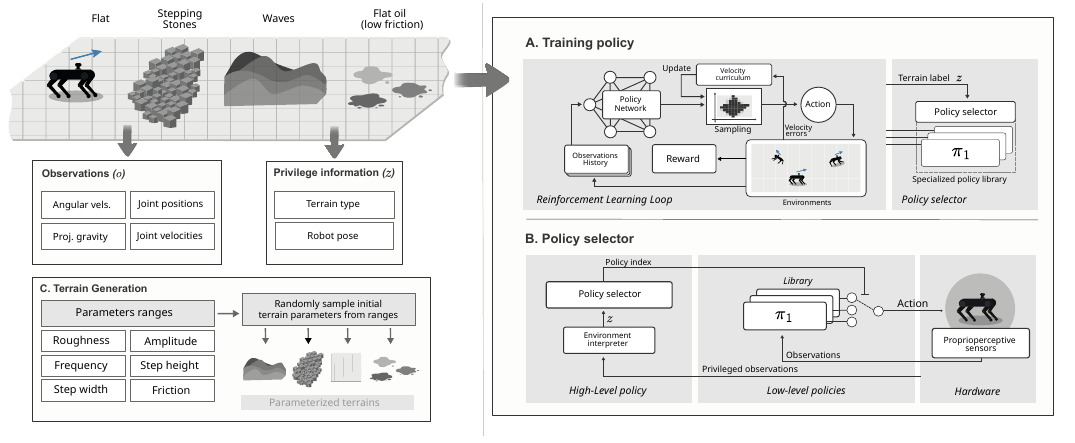}
    \caption{System overview. (A) \textbf{Training Policy}: An RL policy receives a short history of proprioceptive observations, the previous action, and the commanded velocities. The reward blends command-tracking terms with penalties for high kinematic jerk and unstable contacts. A discrete curriculum grid over $(v_x^{\mathrm{cmd}},\ \omega_z^{\mathrm{cmd}})$ is sampled; when a cell's error $\lVert v^{\mathrm{cmd}}-v^{\mathrm{base}}\rVert < \epsilon$, its neighbors unlock, increasing difficulty. Trained policies are appended to a specialized policy library, each stored with its associated privileged observation $z$.
    (B) \textbf{Policy Selector}: A deterministic mapper parses the privileged vector $z$ to extract the relevant cues and directly outputs the index $i$ of the policy to run in the library; the selected $\pi_i$ is applied until a new selection is triggered.
    (C) \textbf{Terrain Generation}: A parametric generator procedurally varies geometric and frictional properties to synthesize a set of challenging terrains.}
    \label{fig:main_diag}
\end{figure*}

\section{Methods}

In this section, we present our method for improving locomotion agility by learning terrain-specialized policies using hierarchical deep reinforcement learning. As illustrated in Fig.~\ref{fig:main_diag}, our approach leverages privileged observations to distinguish between terrain types, enabling the application of terrain-specific curriculum learning strategies adapted to the locomotion limits of each environment. The pipeline includes a terrain generation module that mimics real-world conditions by varying parameters such as surface roughness, amplitude, and friction. After training, a policy selector is used at deployment to activate the policy corresponding to the terrain the robot is currently traversing.

\subsection{Background}

Before presenting our terrain-specialized locomotion approach, we first describe the baseline formulation of a single policy, following standard RL-based control~\cite{lee2020learning}. The task is modeled as a partially observable Markov decision process (POMDP), as not all system states are directly observable under the constraints of blind locomotion. 

To address partial observability, prior work has leveraged belief-state approximations constructed from observation histories. Common strategies include stacking recent observations~\cite{margolis2023walk} or employing long short-term memory (LSTM) architectures to infer latent state information~\cite{han2024lifelike}. In this work, we approximate the POMDP as a fully observable Markov decision process (MDP) by conditioning the policy on a finite history of observations.

Under this approximation, the control problem is defined by the MDP tuple $(\mathcal{S}, \mathcal{A}, \mathcal{T}, \mathcal{R})$, where $\mathcal{S}$ is the set of states $s$, $\mathcal{A}$ the set of actions $a$, $\mathcal{T}(s' \mid s, a)$ the transition function, and $\mathcal{R}(s, a)$ the reward function. The objective is to learn a policy $\pi^*$ that maximizes the expected discounted return:
\begin{equation}
    \pi^* = \arg\max_\pi \mathbb{E}\left[ \sum_{t=0}^{\infty} \gamma^t \mathcal{R}(s, a)_t \right],
\label{eq:policy}
\end{equation}
where, $\gamma \in [0, 1]$ is the discount factor.

Bringing this formulation into the context of legged locomotion, the control problem is handled in discrete time at each timestep $k$. To enable generalization across different legged embodiments, the state is defined in Eq.~(\ref{eq:state}) and is defined solely based on a history of proprioceptive observations, as is standard in blind locomotion.
\begin{equation}
s_k = \{ o_{k - \tau + 1}, \dots, o_k \}, \quad o_i = [ q_i, \dot{q}_i, g_i, v_i, \omega_i]
\label{eq:state}
\end{equation}
where $q, \dot{q} \in \mathbb{R}^{12}$ denote the joint positions and velocities, $g \in \mathbb{R}^3$ is the projected gravity, and $v, \omega \in \mathbb{R}^{3}$ are the linear and angular velocities of the base. The full state $s_k$ thus comprises a history of $\tau$ proprioceptive observations.

Since the objective of this policy is to track locomotion commands while incorporating temporal context, the input state is augmented by concatenating the current command $(v^{\text{cmd}}_x, \omega^{\text{cmd}}_z)$ with the previous action. Additionally, the actions are defined as target joint positions, which are executed using a low-level proportional-derivative (PD) controller that tracks the commanded positions.

To complete the RL formulation, the total reward is defined as the sum of individual terms described in Table~\ref{tab:rw_functions}, comprising components that encourage command tracking and penalize undesirable behaviors such as abrupt joint movements or excessive jumping. The policy is then optimized using the proximal policy optimization (PPO) algorithm~\cite{schulman2017proximal}.

\begin{table}[t]
    \caption{Reward Functions}
    \centering
    \begin{tabular}{l l r}
        \hline
        \textbf{Symbol} & \textbf{Equation} & \textbf{Weight}  \\
        \hline
        $r_{v_{x}}$ & $\exp\left(-( {v}_{x}^{\text{cmd}} - {v}_{x} )^2 / \sigma^2\right)$ & 1.0 \\
        $r_{\omega_z}$ & $\exp\left(-\| \omega_z^{\text{cmd}} - \omega_z \|^2 / \sigma^2\right)$ & 0.5 \\
        $r_{v_{yz}}$ & $\left\| v_{{yz}} \right\|^2$ & $-2.0$ \\
        $r_{\omega_{xy}}$ & $\| {\omega}_{xy} \|^2$ & $-0.05$ \\
        $r_{\tau}$ & $\| {\tau} \|^2$ & $-1.0\mathrm{e}{-5}$ \\
        $r_{\ddot{q}}$ & $\| \ddot{{q}} \|^2$ & $-2.5\mathrm{e}{-7}$ \\
        $r_{\Delta a}$ & $\| {a}_t - {a}_{t-1} \|^2$ & $-0.01$ \\
        $r_{\text{air}}$ & $\sum_{\text{feet}} \mathds{1}_{\text{air}}$ & 0.2 \\
        $r_{\text{contact}}$ & $\sum_{\text{non-foot}} \mathds{1}_{\text{contact}}$ & $-1.0$ \\
        \hline
    \end{tabular}
    \label{tab:rw_functions}
\end{table}

\subsection{Curriculum Learning}

To enable agile locomotion, i.e., the ability to track high-velocity commands, a curriculum learning strategy is necessary. Prior work~\cite{margolis2024rapid} has shown that RL policies struggle to acquire high-speed behaviors when such commands are introduced prematurely during training. Instead, gradually increasing the range of command velocities over time leads to more stable learning and better final performance. Motivated by this insight, we adopt a grid-based curriculum that supports both specialized and generalist locomotion policies, enabling direct comparisons between them.

At each step, a velocity command is sampled from a uniform distribution as defined in Eq.~(\ref{eq:sample_cmd}). The sampling bounds are governed by a discrete curriculum grid, which defines the currently available command space.
\begin{equation}
    v^{\text{cmd}} =
    \begin{bmatrix}
        v_x^{\text{cmd}} \\
        \omega_z^{\text{cmd}}
    \end{bmatrix}
    \sim\!
    \begin{bmatrix}
        \mathcal{U}(v_x^{\text{min}}, v_x^{\text{max}}) \\
        \mathcal{U}(\omega_z^{\text{min}}, \omega_z^{\text{max}})
    \end{bmatrix}
    \label{eq:sample_cmd}
\end{equation}
where $v^{\text{cmd}}$ is the velocity command vector that the robot must track. The subscripts “min” and “max” denote the lower and upper bounds of the sampling interval for each command dimension, which are dynamically adjusted based on the currently unlocked portion of the curriculum grid.

The grid map $\mathcal{G}$ is defined as a discrete matrix of velocity command pairs $(v_x^{\text{cmd}}, \omega_z^{\text{cmd}})$, determined by the maximum command limits and a chosen discretization factor. This grid governs the resolution and coverage of the curriculum space.

The curriculum logic, summarized in Algorithm~\ref{alg:curriculum}, operates by first locating the nearest grid cell associated with the sampled command. The policy’s tracking performance is then evaluated by computing the error between the commanded velocity $v^{\text{cmd}}$ and the robot’s measured velocity $v^{\text{base}}$.  

As shown in Eq.~(\ref{eq:grid_update}), if this error is below a predefined threshold $\epsilon$, the curriculum unlocks the neighboring cells $\mathcal{N}(i,j)$ in the grid, thereby expanding the range of commands the policy can encounter.

\begin{equation}
\mathcal{G}_{k+1} =
\begin{cases}
\mathcal{G}_k \cup \mathcal{N}(i, j), & \text{if } \left\| v^{\text{cmd}} - v^{\text{base}} \right\| < \epsilon \\
\mathcal{G}_k, & \text{otherwise}
\end{cases}
\label{eq:grid_update}
\end{equation}

As new cells are unlocked, the bounds of the sampling distribution are updated by recomputing the minimum and maximum values over the currently available grid, ensuring that the policy gradually progresses toward more challenging commands only after demonstrating reliable tracking performance on simpler ones.

\begin{algorithm}[t]
\caption{Curriculum Update for Velocity Command Grid}
\label{alg:curriculum}
\begin{algorithmic}[1]
\STATE \textbf{Input:} $v^{\text{cmd}} = [v_x^{\text{cmd}},\, \omega_z^{\text{cmd}}]$, $v^{\text{base}} = [v_x,\, \omega_z]$, grid $\mathcal{G}$, threshold $\epsilon$
\vspace{0.5em}
\STATE \textbf{Identify} nearest grid cell to sampled command
\STATE \hspace{1em} $(i, j) \leftarrow f(v_x^{\text{cmd}},\, \omega_z^{\text{cmd}})$
\STATE \textbf{Compute} tracking error
\STATE \hspace{1em} $e \leftarrow \left\| v^{\text{cmd}} - v^{\text{base}} \right\|$
\STATE \textbf{Unlock} neighbors
\IF{$e < \epsilon$}
    \STATE \hspace{1em} $\mathcal{G} \gets \mathcal{G} \cup \mathcal{N}(i, j)$
    \STATE \hspace{1em} $(v_x^{\min}, v_x^{\max}),\, (\omega_z^{\min}, \omega_z^{\max}) \gets \text{bounds}(\mathcal{G})$
\ENDIF
\end{algorithmic}
\end{algorithm}

\subsection{Terrain Generation}
\label{subsec:terrain_gen}
The distribution of training conditions must reflect the variety of surfaces expected at deployment. We address this by using a procedural terrain generator and introducing randomization in physical parameters such as friction coefficients and robot mass. This setup allows the robot to experience diverse contact regimes during training, reducing overfitting to specific terrain types and enabling systematic evaluation of generalization and resilience.

The terrain is sampled according to Eq.~(\ref{eq:terrain_gen}), here a terrain class $c \in \mathcal{C}$ is selected and a corresponding heightmap is generated by $h(x, y)$. The set \(\mathcal{C}\) includes the following terrain types:
\begin{itemize}
    \item {Flat oil:} Planar geometry with reduced friction, modeling slippery floors.
    \item {Waves:} Continuous elevation variation.
    \item {Stepping stones:} Discrete footholds separated by gaps.
    \item {Box grids:} Irregular protrusions and voids.
\end{itemize}

\begin{equation}
\begin{aligned}
&c \sim \mathrm{Cat}(\alpha), \quad c \in \mathcal{C} \\
&h(x, y) = g_c(x, y, \theta_c), \quad \theta_c \sim \mathcal{U}(\theta_c^{\min}, \theta_c^{\max})
\end{aligned}
\label{eq:terrain_gen}
\end{equation}
where $\alpha$ denotes the categorical distribution over terrain classes $c$, $g_c$ is the class-specific heightfield or mesh generator, and $h(x, y)$ represents the resulting surface. The parameter vector $\theta_c$ encapsulates all tunable attributes (e.g., wave amplitude, stone spacing, box height, and friction coefficient).

Each class abstracts a family of real terrains such as slick industrial surfaces, uneven soil, or scattered debris. Parameter sampling within each class supplies intra–class variability, preventing overfitting and enabling systematic evaluation.

\subsection{Terrain-Specialized Policies}

\begin{figure}[htbp]
    \centering

    \begin{subfigure}[b]{\linewidth}
        \centering
        \includegraphics[width=0.56\linewidth]{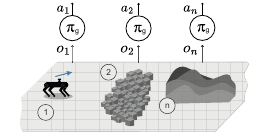}
        \subcaption{Generalist policy}
        \label{fig:generalist}
    \end{subfigure}

    \vspace{0.6em}

    \begin{subfigure}[b]{\linewidth}
        \centering
        \includegraphics[width=0.56\linewidth]{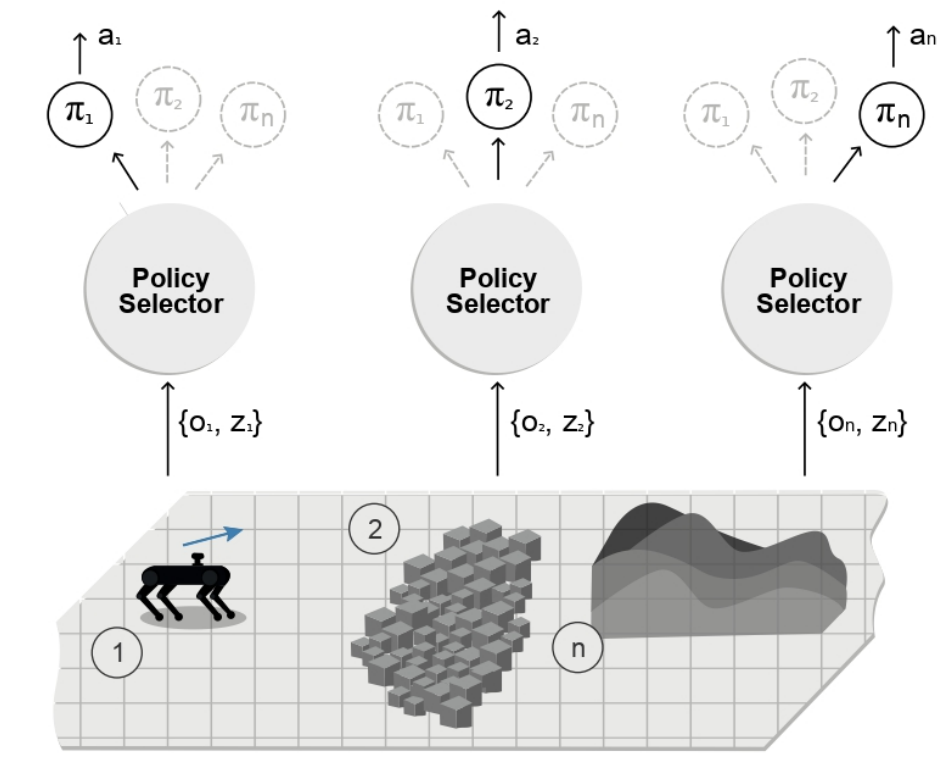}
        \subcaption{Specialist policy}
        \label{fig:specialist}
    \end{subfigure}

    \caption{Comparison of a) Generalist and b) Specialist policy.}
    \label{fig:policy_comparison}
\end{figure}

Building on the previously defined MDP formulation and curriculum strategy, we now describe our approach for training terrain-specialized policies. As illustrated in Fig.~\ref{fig:generalist}, we define a generalist policy as a single policy trained to operate across all terrain conditions. This policy receives the observation vector defined in Eq.~(\ref{eq:state}) and is trained to output the best action regardless of the specific terrain. However, due to the challenges posed by different terrains---such as varying friction, deformability, or geometry---a single policy may struggle to learn optimal behaviors across all conditions, reducing agility and using suboptimal motion patterns.

To address this, we propose learning terrain-specialized policies, as shown in Fig.~\ref{fig:specialist}. In this setup, a separate policy is trained for each terrain type, allowing it to specialize in the locomotion strategies most effective under those conditions. This approach introduces an additional input signal, denoted by $z$, which serves as a privileged observation indicating the current terrain. This information is used by a policy selector module to route training data to the appropriate policy and, during deployment, to execute the corresponding policy.

This hierarchical setup allows each specialized policy to focus exclusively on a single terrain, enhancing its ability to learn terrain-specific agility and motion patterns. When combined with the curriculum learning strategy, each specialized policy can explore a broader range of velocity commands and effectively learn to track them. This is possible because the robot is not constrained by uncertainty about the current environment; instead, the privileged terrain information enables targeted learning within a known context.

\newpage

\section{Experimental Results}

In this section we present the experimental results of our hierarchical locomotion framework using terrain-specialized policies. We begin by describing the experimental setup, which includes the hardware platform, software tools and simulation environment used for evaluation. We then compare the performance of the policy selector, which chooses the most suitable specialized controller during execution, against a single generalist policy. The comparison is carried out through two evaluations. The first consists of a dense sweep of velocity commands on isolated terrains. The second uses a continuous track composed of thirteen interleaved terrain segments. Success rate, tracking error and traversal time are measured to assess overall performance.

\subsection{Platform}

For our experiments, we used the ANYmal-D robot (ANYbotics), a quadrupedal platform with a symmetric design and proven robustness for traversing rough and challenging terrains. All training and evaluation were conducted entirely in simulation using IsaacSim~\cite{mittal2023orbit}, chosen for its high-fidelity physics engine, support for parallelized environments with GPU acceleration, and native integration with RL libraries. 

\subsection{Policy Training}

\begin{table}[b]
    \caption{PPO Training Hyperparameters}
    \centering
    \small
    \begin{tabular}{l l}
        \hline
        \textbf{Parameter} & \textbf{Value} \\
        \hline
        Clip ratio ($\epsilon$) & 0.2 \\
        Learning rate & $10^{-3}$ \\
        Discount factor ($\gamma$) & 0.99 \\
        GAE-lambda & 0.95 \\
        Desired KL divergence & 0.01 \\
        Entropy coefficient & 0.005 \\
        Value loss coefficient & 1.0 \\
        Batch size & 36  $\cdot$ 4096  \\
        Actor hidden dims & [256, 256, 128] \\
        Critic hidden dims & [256, 256, 128] \\
        \hline
    \end{tabular}
    \label{tab:ppo_parameters}
\end{table}

To train our policies on the terrain configurations described in Subsection~\ref{subsec:terrain_gen}, we used the \texttt{rsl-rl} library~\cite{rudin2022learning}, employing the PPO algorithm with the hyperparameters detailed in Table~\ref{tab:ppo_parameters}. The same training configuration was uniformly applied to the generalist policy as well as the specialized policies; this included a historical observation window of $\tau = 15$ and a curriculum learning scheme based on velocity commands sampled from a range of $[-3.0, 3.0]$, discretized into $625$ grid cells. Progression within the curriculum was governed by tracking error thresholds of $\epsilon_{v_x} = 0.15$ for linear velocity and $\epsilon_{\omega_z} = 0.25$ for angular velocity.

\begin{figure}[h]
    \centering
    \includegraphics[width=0.9\linewidth]{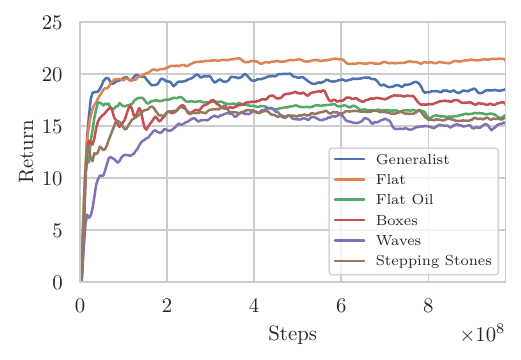}
    \caption{Training curves of average return against simulation steps: generalist vs. five specialized policies.}
    \label{fig:rw_policies}
\end{figure}

\begin{figure}[h]
    \centering
    \includegraphics[width=0.9\linewidth]{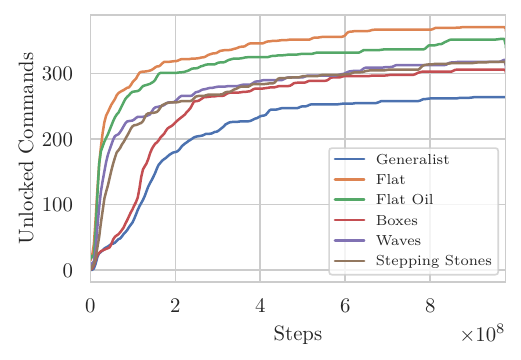}
    \caption{Unlocked velocity commands refer to those the robot successfully tracked with error below the threshold $\epsilon$.}
    \label{fig:curriculum_policies}
\end{figure}

The training progress of the policies described above is shown in Fig.~\ref{fig:rw_policies}. While all policies demonstrate convergence, it is important to note that the curriculum challenge is not explicitly rewarded. Therefore, the return curves alone are not sufficient to assess how well each policy generalizes across different command velocities. For this reason, Fig.~\ref{fig:curriculum_policies} illustrates the velocity commands that were unlocked during curriculum learning. These correspond to commands the robot successfully tracked with an error below the defined threshold $\epsilon$, thereby enabling expansion to neighboring cells.

By combining both results, we can conclude that although all policies exhibited convergence---indicating that a locomotion behavior was learned---the specialist policies unlocked a greater number of velocity commands. This suggests that specialist policies are more robust across a wider range of commands compared to the generalist. 

\subsection{Specialized Locomotion Validation}

\begin{table*}[t]
    \caption{Performance Comparison between Generalist and Specialized Policies on Continuous Multi-Terrain}
    \centering
    \begin{tabular}{cccccc}
        \toprule
        \textbf{Linear Vel. [m/s]} & \textbf{Model} & \textbf{Success Rate [\%]} & \textbf{Tracking Error [$\sqrt{\text{m/s}}$]} & \textbf{Total Distance [m]} & \textbf{Fail Distance [m]} \\
        \midrule
        \multirow{2}{*}{0.5} & Generalist  & 76.0 & 0.165 $\pm$ 0.194 & 53.14 $\pm$ 12.71 & 31.43 $\pm$ 7.28\phantom{0} \\
                             & Specialized & \textbf{96.0} & \textbf{0.053} $\pm$ 0.091 & \textbf{58.58} $\pm$ 6.95\phantom{0}  & 24.53 $\pm$ 0.00\phantom{0} \\

        \multirow{2}{*}{1.0} & Generalist  & \textbf{92.0} & \textbf{0.123} $\pm$ 0.259 & \textbf{58.55} $\pm$ 4.91\phantom{0}  & 41.92 $\pm$ 0.15\phantom{0} \\
                             & Specialized & 88.0 & 0.167 $\pm$ 0.307 & 54.94 $\pm$ 13.71  & 17.83 $\pm$ 0.00\phantom{0} \\

        \multirow{2}{*}{1.5} & Generalist  & \textbf{96.0} & \textbf{0.126} $\pm$ 0.281 & \textbf{58.53} $\pm$ 7.19\phantom{0}  & 23.31 $\pm$ 0.00\phantom{0} \\
                             & Specialized & 84.0 & 0.301 $\pm$ 0.523 & 54.62 $\pm$ 12.32  & 26.80 $\pm$ 1.09\phantom{0} \\

        \multirow{2}{*}{2.0} & Generalist  & 40.0 & 1.256 $\pm$ 0.912 & \textbf{55.42} $\pm$ 7.10\phantom{0}   & 54.91 $\pm$ 3.27\phantom{0} \\
                             & Specialized & \textbf{76.0} & \textbf{0.561} $\pm$ 0.809 & 54.55 $\pm$ 11.44 & 41.44 $\pm$ 8.94\phantom{0} \\

        \multirow{2}{*}{2.5} & Generalist  & 4.0\phantom{0} & 2.411 $\pm$ 0.436 & \textbf{42.89} $\pm$ 13.44  & 45.48 $\pm$ 11.04 \\
                             & Specialized & \textbf{44.0} & \textbf{1.469} $\pm$ 1.163 & 39.43 $\pm$ 22.69  & 46.69 $\pm$ 17.56 \\
                             
        \bottomrule
    \end{tabular}
    \label{tab:multi_terrain_comparison}
\end{table*}

The first validation experiment evaluated all trained policies across a predefined range of velocity commands, with $v_x \in [-5.0, 5.0]$ and $\omega_z \in [-5.0, 5.0]$, discretized into 41 intervals, resulting in 1681 unique velocity command pairs. For each pair, we computed the tracking error between the commanded and measured velocities over 5 trials of 900 steps, discarding the first 50 steps as warm-up.

Table~\ref{tab:terrain_thresholds} summarizes the results. This evaluation was conducted independently for each specialized terrain, allowing a direct comparison between the specialized and generalist policies. Tracking performance was assessed under varying thresholds $\delta$, which defined the criterion for successful command execution.

Analyzing the results, we observe that specialist policies consistently outperform the generalist in tracking a broader range of velocity commands. This underscores the importance of using terrain-specialized policies to achieve agile locomotion in blind settings, especially at higher speeds. The performance gap depends on the complexity of the environment. For instance ,the most significant improvement is observed on the flat oil terrain, where leg slippage increases the difficulty of control, as shown in Table~\ref{tab:terrain_thresholds}. In contrast, the flat terrain presents a simpler scenario in which both specialist and generalist policies perform similarly, since the difference in tracked commands at $\delta = 0.50$ is less than 5\%.

\subsection{Continuous Multi-Terrain Evaluation}

The second experiment aimed to compare the performance of the hierarchical locomotion control architecture, based on a policy selector, with that of the generalist policy. In this setup, a privileged observation variable $z$ was extracted from the simulator to identify the current terrain type. This information was used to switch between specialized policies.

\begin{table}[b]
    \centering
    \caption{Tracking Performance of Locomotion Policies}
    \label{tab:terrain_thresholds}
    \begin{threeparttable}
    \footnotesize
    \begin{tabular}{llcc}
        \toprule
        \textbf{Terrain} & \textbf{Model} & \multicolumn{2}{c}{\textbf{Tracking Error Threshold ($\delta$)}} \\
        \cmidrule(lr){3-4}
        & & \phantom{000}0.25 & \phantom{00}0.50 \\
        \midrule
        \multirow{2}{*}{Flat}
            & Generalist & \phantom{0000}80.4\% & \phantom{000}93.1\% \\
            & Specialist & \phantom{0000}\textbf{90.1\%} & \phantom{000}\textbf{94.7\%} \\
        \midrule
        \multirow{2}{*}{Flat Oil}
            & Generalist & \phantom{0000}31.9\% & \phantom{000}47.1\% \\
            & Specialist & \phantom{0000}\textbf{65.9\%} & \phantom{000}\textbf{77.9\%} \\
        \midrule
        \multirow{2}{*}{Waves}
            & Generalist & \phantom{0000}47.5\% & \phantom{000}66.8\% \\
            & Specialist & \phantom{0000}\textbf{58.3\%} & \phantom{000}\textbf{76.2\%} \\
        \midrule
        \multirow{2}{*}{Boxes}
            & Generalist & \phantom{0000}54.0\% & \phantom{000}76.6\% \\
            & Specialist & \phantom{0000}\textbf{69.0\%} & \phantom{000}\textbf{84.7\%} \\
        \midrule
        \multirow{2}{*}{S. Stones}
            & Generalist & \phantom{0000}66.7\% & \phantom{000}84.3\% \\
            & Specialist & \phantom{0000}\textbf{76.4\%} & \phantom{000}\textbf{90.6\%} \\
        \bottomrule
    \end{tabular}
    \vspace{3pt}
    \begin{tablenotes}\footnotesize
    \item[*] Each cell reports the average success rate for tracking \textbf{linear} ($v_x$) and \textbf{angular} ($\omega_z$) velocity commands.
    \item[*] The generalist policy was evaluated across all terrains, while each specialist was evaluated only on its corresponding terrain.
    \end{tablenotes}
    \end{threeparttable}
\end{table}

To evaluate this approach, we constructed a continuous multi-terrain track of 65 meters, consisting of segments from different terrain types. Both the hierarchical controller and the generalist policy were tasked with traversing the entire sequence under the same initial conditions and command profiles, with a constant linear velocity in each trial. A proportional controller was used to generate angular velocity commands to maintain the robot’s alignment with the center of the path. The track included flat terrain, uneven surfaces with varying amplitudes, discrete obstacles such as boxes and stepping stones, and low-friction patches, arranged in a challenging sequence. Failures to track were treated as full command errors when computing the velocity tracking metrics, as summarized in Table~\ref{tab:multi_terrain_comparison}.

\begin{figure}[b]
    \centering
    \includegraphics[width=1.0\linewidth]{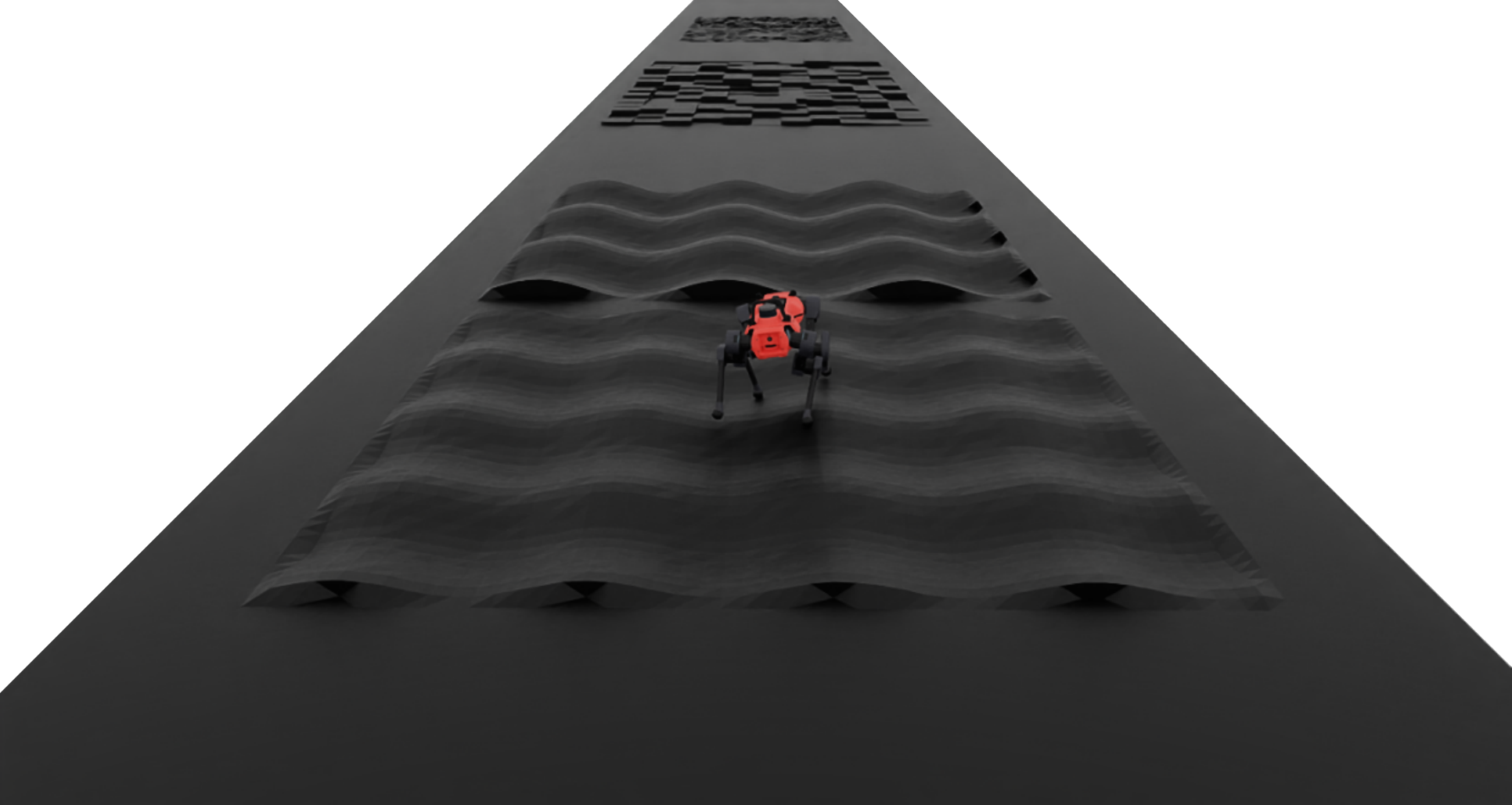}
    \caption{Illustrative simulation image showing the sequential multi-terrain evaluation track.}
    \label{fig:long_terrain}
\end{figure}

\begin{figure}[h]
    \centering
    \includegraphics[width=0.92\linewidth]{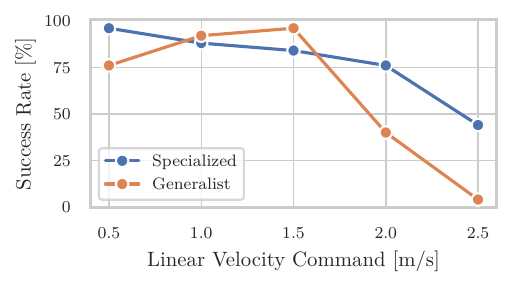}
    \caption{Traversal success across mixed terrains under progressively higher commanded speeds.}
    \label{fig:success_compare}
\end{figure}

\begin{figure}[h]
    \centering
    \includegraphics[width=0.92\linewidth]{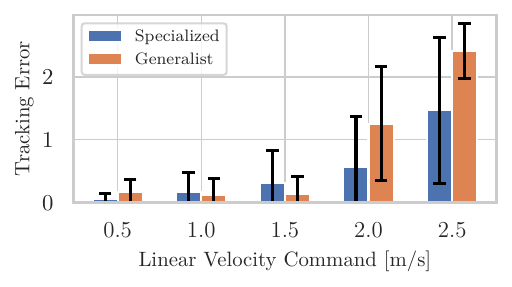}
    \caption{Mean tracking error along multiterrain course for varying speed demands.}
    \label{fig:tracking_error}
\end{figure}

The mixed-terrain track reveals a consistent performance gap between the controllers. As shown in Fig.~\ref{fig:success_compare} and Fig.~\ref{fig:tracking_error}, at lower commanded speeds, both the generalist policy and the specialized-policy complete the course with comparable success rates and minimal tracking errors. This is likely because friction and impact dynamics remain within the bounds covered by domain randomization during training, making terrain-specific specialization less impactful. However, as the target speed increases, more abrupt contact transitions and reduced traction lead the generalist to accumulate slip-induced errors and fail more runs. Furthermore,  across all trials, the generalist policy achieved a 61.6\% success rate, lower than the 77.6\% observed with the specialized policies.

In contrast, the selector activates the policy specialized for the prevailing surface, resulting in lower trajectory deviation and a higher rate of successful completions. While both approaches experience performance degradation at higher speeds, the specialist ensemble exhibits a slower decline in performance. These results highlight its superior robustness and reliability under challenging operating conditions.

\section{Conclusion}

We introduced a hierarchical reinforcement learning framework for blind legged locomotion, combining terrain-specialized policies with curriculum learning to enhance agility and robustness. By decomposing the task and training specialized controllers, our method improves tracking accuracy and adaptability across diverse and challenging terrains. Simulation results demonstrate the advantages of specialization, particularly under high-speed and low-friction conditions. Future work will explore techniques to eliminate reliance on privileged information during deployment and to enable sim-to-real transfer on the physical robot.
\section*{Acknowledgment}

This work was supported by São Paulo Research Foundation (FAPESP) grant no 2025/04308-7 and Petrobras, using resources from the R\&D clause of the ANP, in partnership with the University of São Paulo and the intervening foundation FAFQ, under Cooperation Agreement No. 2023/00016-6 and 2023/00013-7, as well as by the Brazilian National Research Council (CNPq) grants no. 308092/2020-1.

\addtolength{\textheight}{-12cm}   



\bibliographystyle{IEEEtran}
\bibliography{bibtex}

\end{document}